\documentclass{INTERSPEECH2023}

\interspeechcameraready

\usepackage{tabularx}

\title{Tri-level Joint Natural Language Understanding \\ for Multi-turn Conversational Datasets}

\name{Henry Weld$^1$, Sijia Hu$^1$, Siqu Long$^1$, Josiah Poon$^1$, Soyeon Caren Han$^{1,2,*}$}

\address{
  $^1$The University Of Sydney, Australia,
  $^2$The University Of Western Australia, Australia
  $^*$Corresponding Author}
\email{hwel4188@uni.sydney.edu.au, sihu9591@uni.sydney.edu.au, slon6753@uni.sydney.edu.au, Josiah.Poon@sydney.edu.au, Caren.Han@uwa.edu.au}

\usepackage[outline]{contour}

\usepackage[stable]{footmisc}
\begin{document}

\maketitle
\begin{abstract} 
Natural language understanding typically maps single utterances to a dual level semantic frame, sentence level intent and slot labels at the word level. The best performing models force explicit interaction between intent detection and slot filling. We present a novel tri-level joint natural language understanding approach, adding domain, and explicitly exchange semantic information between all levels. This approach enables the use of multi-turn datasets which are a more natural conversational environment than single utterance. We evaluate our model on two multi-turn datasets for which we are the first to conduct joint slot-filling and intent detection. Our model outperforms state-of-the-art joint models in slot filling and intent detection on multi-turn data sets. We provide an analysis of explicit interaction locations between the layers. We conclude that including domain information improves model performance\footnote{A circuit diagram and associated code is available at https://github.com/adlnlp/Tri-NLU}.
\end{abstract}

\section{Introduction} \label{chap:nlu process}
Natural Language Understanding (NLU) is a field of natural language processing (NLP) that studies how computers understand natural human language. The concept of a hierarchical semantic frame of domain-intent-slot has emerged to represent the meaning of natural language. For each spoken utterance, the \textit{domain} is the conversational topic, e.g. \textit{restaurant}. \textit{Intent} represents the purpose of the user's current utterance, like naming the restaurant or confirming a reservation. A token level labelling describes the semantic role of each word in the sentence, termed the \textit{slots} \cite{tur2011spoken}. The tasks comprising the filling of the frame are called domain classification (DC), intent detection (ID) and slot filling (SF). SF is a sequence labelling task, whereas ID and DC are classification tasks on the entire utterance. In Table \ref{tab:sf-id-example}, we have the utterance (or turn) ``Play the movie RMS Titanic", and its annotations for intent, domain, and slots. The slot annotation uses the Beginning-Outside-Inside (BOI) format. The ``movie" token is the entertainment type, and the span of tokens ``RMS Titanic" is the movie's title. An NLU circuit should accurately determine that the user intends the dialogue system to play a movie and that the movie name is ``RMS Titanic". Or, the NLU component classifies the user's intent and finds the relevant slot-value pairs. The domain here is entertainment; it is from a multi-domain personal assistant dataset. 

Most existing NLU datasets only include single-turn utterances in a single domain, and so a large proportion of the literature focuses on determining how to categorise a single utterance's intent and slots \cite{weld2022survey}. However, several recent multi-turn datasets allow annotation at the dialogue level and across multiple domains \cite{39,budzianowski2018multiwoz, zang2020multiwoz, weld2021conda}. Dialogue state tracking systems widely use these multi-turn datasets. When given both the current utterance and the dialogue history, they determine whether a specific slot-value pair is mentioned in the current utterance \cite{80}.

\begin{table}[t]
\caption{An example sentence with its intent, domain and slot-value pairs annotated}
\label{tab:sf-id-example}
\centering
\begin{tabular}{|l|lllll|}
\hline
\textbf{Utterance} &
  \multicolumn{1}{l|}{Play} &
  \multicolumn{1}{l|}{the} &
  \multicolumn{1}{l|}{movie} &
  \multicolumn{1}{l|}{RMS} &
  \multicolumn{1}{l|}{Titanic} \\ \hline
\textbf{Slot} &
  \multicolumn{1}{l|}{O} &
  \multicolumn{1}{l|}{O} &
  \multicolumn{1}{l|}{B-ent\_type} &
  \multicolumn{1}{l|}{B-title} &
  \multicolumn{1}{l|}{I-title} \\ \hline
\textbf{Intent}   &  \multicolumn{4}{l}{play\_movie} &   \\ \hline
\textbf{Domain}   &    \multicolumn{4}{l}{entertainment}    & \\ \hline
\end{tabular}
\end{table}

\subsection{Related work}\label{chap:litreview}

Many researchers since 2013 have focused on presenting joint models, which process information from both slot and intent processing simultaneously to better capture the joint distributions of intent and slot labels in the utterance \cite{weld2022survey}. Earlier models used an implicit sharing of information between the tasks, typically in the form of a common encoder, and then a joint loss after parallel paths through the network concentrating on ID and SF respectively \cite{zhang2016joint, liu2016attention}. 
Better performance was exhibited when the sharing of information between the tasks is explicit. In \cite{goo2018slot} a weighted sum of an RNN's hidden states is calculated at each time step and serves as the input for both the SF and ID tasks. An attention mechanism produces a slot context vector at each time step and a global intent context vector incorporating the sequence-level information. The paper proposes a ``slot gate." At each time step, the slot gate is a weighted sum of the slot context vector and the intent context vector. The slot gate informs slot prediction and is an early example of intent2slot, where the parameters from the intent prediction are also directly provided to each slot prediction at each time step. Intent is only informed by the intent context vector so there is no slot2intent aspect. Slot2Intent was then added by \cite{e2019novel} who proposed a model with an additional intent gate, which injects contextual information from the slot-filling task into intent detection. BERT was introduced in NLU for joint modeling by simply taking the BERT encodings of the utterance to perform softmax predictions on each token (including [CLS] for intent) and calculating a joint loss \cite{chen2019bert}. Still, it was immediately effective due to the Transformer \cite{vaswani2017attention} architecture capturing coincidence of words as more important in NLU than the temporal order of arrival emphasised by RNN models. Other successful models used BERT in combination with explicit sharing of information between tasks \cite{qin2019stack, zhang2019joint}.
\cite{han2021bidir} used explicit intent2slot and slot2intent influence for their joint model, which uses an intent probability distribution paired with BERT representations of the tokens in the slot-filling task and uses the slot label probability distribution along with BERT's [CLS] token representation for intent detection. 

Separate encodings for slot and intent representations at the start of NLU circuits may enable different learned focus on the data for different task purposes. \cite{hui2021clim} used a BiLSTM with its stronger positional information for slot filling and a Transformer for intent semantics. The separate slot and intent representations later interact via attention. In \cite{qin2021coint} a pair of Transformer encoders act on Bi-LSTM encodings of Glove word vectors. Within the encoders some exchange of information takes place, via the query vectors, self attention output, and via a common Feed Forward layer. While the model performs well on the standard joint task benchmark datasets the ablations performed could not explain which parts of the exchange of information are contributing most.

There are several existing multi-turn datasets for dialogue state tracking which with some pre-processing can be used for joint NLU. One approach is to interpret slot, intent and domain as a hierarchy in the semantic frame thus produced. \cite{hakkani2016multi} used a private multi-domain data set and addressed the task by combining domain and intent labels into one tag. In \cite{kim2017onenet} domain is classified at the level of the utterance using a simple attention mechanism over RNN hidden states. \cite{firdaus2020deep} use dialogue action as a domain proxy and detect it using a similar mechanism.

\subsection{The Contribution}
In NLU the standard datasets are single utterance annotated to a dual level semantic frame. Performance on these benchmarks is saturated \cite{weld2022survey}. In this paper,  we extend the frame and explore whether domain representations can then provide an additional layer of context information to both SF and ID tasks, as the semantics of an utterance are highly dependent on its domain. The purpose of this research is to explicitly incorporate domain information into the joint model of SF and ID in order to improve the performance of both tasks on multi-turn datasets. We note that we do not track dialogue history. We use each utterance's intent, slot, and domain annotation in the same way we handle single-turn datasets, albeit with an extra semantic level. 

\begin{itemize}
    \item We extend the use in NLU of multi-turn, tri-level semantic frame datasets in slot filling and intent detection experiments.
    \item We propose a novel tri-joint intent-slot-domain model that explicitly shares information among slot filling, intent detection, and domain classification. We show the inclusion of a domain increases performance in the intent and slot tasks; 
    \item We investigate the location of information exchange between Transformer encoders aligned to each task to best improve ID and SF performance.
\end{itemize}

\section{Methodology}\label{chap:methodology}
Our model architecture consists of three layers: the input embedding layer, the tri-Transformer-encoder layer, and the tri-directional joint NLU layer. 



\subsection{Input embedding layer}\label{chap:method:bert}
Three separate BERT\footnote{BERT-base-uncased model: https://pypi.org/project/pytorch-pretrained-BERT/} encoders are trained on intent, domain classification and slot filling tasks respectively. Each encoder produces representations of dimension $d = 768$ for each token in the input sentences, including [CLS] and [SEP] tokens. Our sequence length is set to 20 for performance purposes. 

\subsection{Tri-Transformer-Encoder Layer}
On top of each BERT instance in the input embedding layer we place a two-layer Transformer encoder, to further process one each of the intent detection, slot filling, and domain input embedding data. Each single-layer Transformer encoder contains 4 successive sub-layers. 1) a multi-head self-attention (SA) layer which performs the scaled Key-Query-Value Attention. 3 heads are used. 2) A first Add \& Norm layer which reconstructs the SA outputs back to dimension $d$, adds it to the initial input representations and performs normalisation 3) A position-wise, fully connected feed-forward (FF) layer. 4) A second Add \& Norm layer which adds the output of the FF layer to its input and normalises. All model embedding and sub-layers provide outputs of size $d$. To produce the key (K), query (Q) and value (V) inputs to each Transformer encoder the output of its input embedding layer is passed through separate linear layers, one for each of K, Q and V. 
Three variants are experimented with: non-exchange, cross-attention and before-feed-forward. In the non-exchange (NoEx) variant there is no information exchange between the Transformer encoders. Thus, slot filling, intent detection, and domain classification use distinct Transformer encoders. The other variants allow interaction between the slot encoder and the intent and domain encoders. Interaction between intent and domain occurs later in the circuit.

\subsubsection{Cross-Attention Variant}

\begin{figure}
\begin{center}
\includegraphics[width=3in]{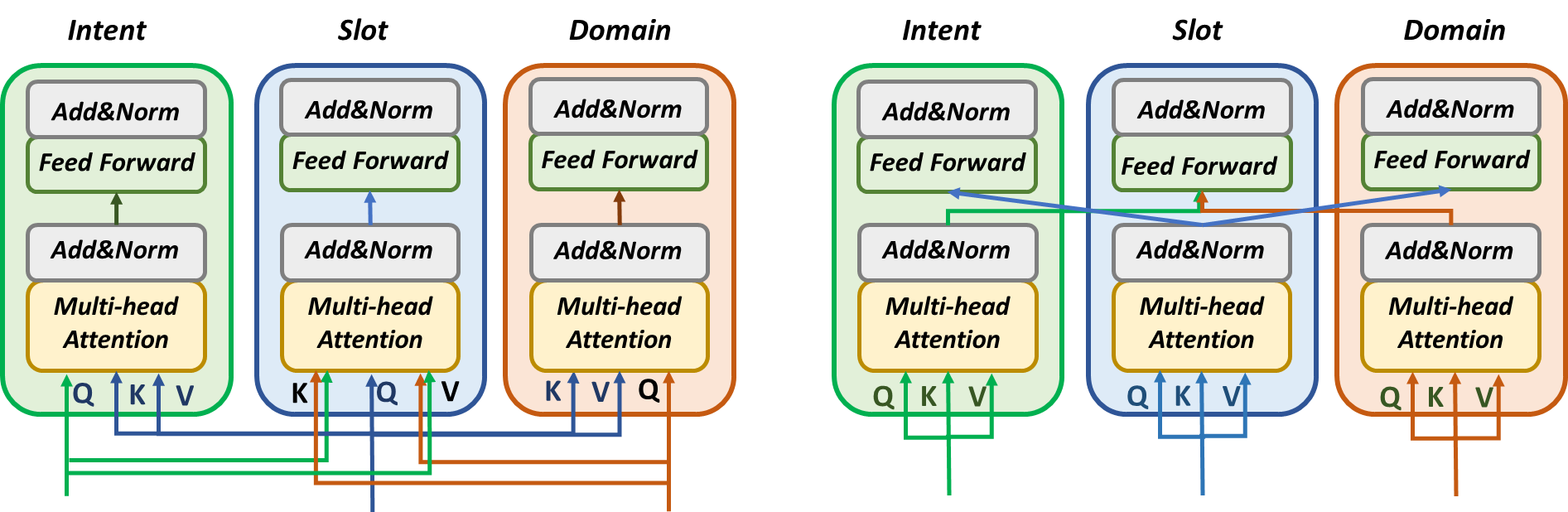}
\caption{Architecture of the Cross-Attention (left) and before-feed-forward (right) Tri-Transformer-Encoder variants}
\label{fig:bf}
\end{center}
\end{figure}

We introduce explicit information exchange between encoders. In Figure \ref{fig:bf} (left) the slot encoder is in the middle. 
Each SA layer takes its own Q. The left (intent) and right (domain) SA layers take the K and V from the middle encoder. The middle SA layer concatenates the K of the left (intent) encoder and right (slot) encoder then projects down to size $d$ for its K input. Similarly it concatenates the V of the other two encoders then projects down to size $d$ for its V input. The procedure afterwards is the same as the NoEx variant. 




\subsubsection{Before-Feed-Forward Variant}


The before-feed-forward variant (BF) (Figure \ref{fig:bf} (right)) feeds the result of the SA layer of the middle (slot) encoder as the input of the FF layer of the left (intent) and right (domain) encoder FF layers. The middle unit concatenates the SA outputs of the left and right encoders then projects down to size $d$ for its FF input. 
The three outputs from any of the variants of the tri-Transformer-encoder layer are $H^S=(h^S_1,..., h^S_n)$, $H^I=(h^I_1,..., h^I_n)$ and $H^D=(h^D_1,..., h^D_n)$, each tuple containing vectors of size $d$ for each token in the input sequence (the BERT CLS and SEP tokens are not used).

\subsection{Tri-directional Joint NLU Layer}
Fusion and prediction take place in a tri-directional Joint NLU layer on top of the tri-Transformer-encoder layer, which involves direct incorporation of information from the other two tasks into each task. Let $n_s$, $n_i$, and $n_d$ represent the number of distinct slot, intent, and domain labels.

$H^I$ is flattened (the elements concatenated), has a tanh activation applied, and is passed through a linear layer to size $n_i$ upon which a softmax is performed to give a probability distribution over the intent classes.
\begin{align}
{H^I}' &= tanh(flatten(H^I)) \\
P_{I} &= softmax({H^I}' W_{I} + b_{I})
\end{align}

Similarly we produce a probability distribution over the domain classes, while for the initial slot probability distribution, we produce a probability distribution over the slot labels for each token:
\begin{align}
{H^D}' &= tanh(flatten(H^D)) 
\\ P_{D} &= softmax({H^D}'  W_{D} + b_{D})
\end{align}



SF takes the semantic information from the whole sequence at both the intent and domain levels to predict a slot label for each word. We make a copy of the $P_{I}$ and $P_{D}$ for each token then concatenate them with the hidden states from the slot Transformer encoder $H^S$. These are then projected again to vectors of length $n_s$ and a softmax produces slot label probability distributions for each token which can be fed to the slot loss function. With ${P_{I}}'$ and ${P_{D}}'$ the repeated copies of the probability vectors:

\begin{align}
S_{slot} &= (H^S  \oplus {P_{I}}' \oplus {P_{D}}') \\
P_{S} &= softmax(S_{slot} \times W_{slot} + b_{slot}) 
\end{align}

For ID and DC we use the slot sequence embeddings to give more context. To map the slot sequence information to the same dimension as intent and domain labels, we first flatten the slot Transformer encoder output $H^S$ then concatenate it with the intent and domain information and project to size $n_i$ and $n_d$ respectively then perform softmaxes to provide inputs to the intent and domain loss functions:
\begin{align}
S &= flatten(H^S) \oplus {H^I}' \oplus {H^D}' \\
I &= softmax(S  W^{concat}_{I} + b^{concat}_{I}) \\
D &= softmax(S  W^{concat}_{D} + b^{concat}_{D}) 
\end{align}

The tri-Joint NLU model is trained on the sum of the cross entropy losses for slot filling, intent detection, and domain classification.

\section{Experiment}\label{chap:experiment}
Our proposed model performs three NLU tasks simultaneously. Given an input utterance of tokens $X = (x_1, x_2, ..., x_T)$, the objective of our task is to produce a mapping:
    \[NLU(X) = (Y^{slot}, y^{intent}, y^{domain}) \]

\noindent where $y^{intent}$ and $y^{domain}$ are the singular intent and domain labels of the utterance and $Y^{slot}=(y^{slot}_1, y^{slot}_2, ..., y^{slot}_{T})$ is a tuple of slot labels positionally aligned with the input tokens. We focus on the inclusion of the domain task on intent detection and slot filling, and we use the metrics from NLU of intent accuracy and token based slot F1, with higher scores indicating superior performance. 
\subsection{Implementation Details}
All results are quoted for 20 epochs. The model has 29 million parameters. Training time is 6 minutes for M2M and 19 minutes for MultiWOZ on Google Colab GPU. Default initialisation is used. Grid search is used for hyperparameter tuning with only the final model described herein.

\subsection{Multi-turn NLU Datasets}\label{chap:multiturn nlu}
We use two English multi-turn datasets which are used as benchmarks in the field of dialogue state tracking (DST). 

\textbf{Machines Talking To Machines (M2M)} multi-turn dataset\footnote{M2M Dataset: https://github.com/google-research-datasets/simulated-dialogue} uses virtual agent and virtual user-generated interactions to replicate how people talk to each other in a goal oriented dialogue. The M2M dataset covers two domains: restaurant and movie. A dialogue, of average length 9.86 turns, is within one domain. The average tokens per turn is 8.24. Each turn (or utterance) is annotated with a dialogue act which we use as a proxy for intent. There are 15 intents (e.g. \textit{greeting, inform, confirm, request, affirm, negate, notify\_success)}, 6 restaurant slots (\textit{price\_range, location, rest\_name,category, num\_people, date, time}) and 5 movie slots (\textit{theatre\_name, movie, date, time, num\_people}). There are 13974 training and 7998 test samples.


\textbf{Multi-Domain Wizard-of-Oz (MultiWOZ) 2.2}\footnote{MultiWOZ 2.2: https://github.com/budzianowski/multiwoz} is a multi-turn dataset which is designed for dialogue state tracking (DST) tasks \cite{zang2020multiwoz}. It aims to record a realistic conversation between a tourist and a person working at city tourism information desk. It is annotated for domain, intent and slot. Multi-labels for domain and intent are combined into a single label. There are 48 domains, 78 intents and 17 slots in 47897 training samples and 6251 test samples. The dialogues and turns are longer than M2M, at 13.46 turns per dialogue and 13.13 tokens per turn.

\subsection{Baseline Models}
Our baselines are the pre-trained neural network language models $BERT_{base}$ and $BERT_{large}$ \cite{devlin2019bert}, and 
$ALBERT_{base}$ and $ALBERT_{large}$ \cite{83}, used in the JointBERT fashion of \cite{chen2019bert}. Then, in a separate test, we add an 2 layer transformer encoder with no exchange to the pre-trained models. We also pass the data through three benchmark joint intent detection and slot filling architectures, performing just the SF and ID tasks. We use the Sequence-to-sequence joint model \textbf{(Seq2Seq)} of \cite{liu2016attention} and the Slot-Gated joint model \textbf{(Slot-Gated)} of \cite{goo2018slot}. The third model is Bi-LSTM encoder-decoder with stacked CRF \cite{li2019joint}\textbf{(Bi-LSTM + CRF)}. The last encoder hidden state is used for ID, and as an input to the slot sequence decoder which leverages an attention unit to learn context vectors for decoder states. The stacked CRF layer then forecasts slot labels after receiving the outputs of the LSTM decoder \cite{85}. We are the first to test these joint SF and ID models on multi-turn dialogue scenarios. 

\section{Evaluation Results}
\subsection{Overall Performance\footnote{For Table \ref{tab:Bert m2m} and \ref{tab:joint multiwoz}, note that we used Cross-Attention Variant interaction on M2M, and BF Variant interaction on MultiWOZ}}

We first compare the evaluation of our model the best Slot F1 performance to different variants of pre-trained BERT. We performed five-fold cross-validation on M2M and MultiWOZ dataset, and the average of the experiments is reported in Table \ref{tab:Bert m2m}. On both datasets, our model with $BERT_{base}$ gives the best results compared to eight BERT variants. Compared to the best baseline model, $BERT_{large}SA$, our model with Cross-Attention variant interaction achieves 0.13\% higher Intent Detection (ID) accuracy and 2.46\% higher Slot Filling (SF) F1 on the M2M dataset, and our model with BF variant interaction scores 3.48\% higher ID accuracy and 0.24\% higher SF F1 on the MultiWOZ dataset. We conclude that introducing the explicit interaction structure as in our model, rather than increasing the model size, gives better performance on the NLU joint task. While the extra independent Transformer encoder layers improve each JointBERT model, the extra interaction opportunities in our model give locations to better learn the joint distribution of intent, domain and slot labels.

\begin{table}
\caption{Comparison results with BERT variations. SA indicates inclusion of extra Transformer encoder layers.} 
\label{tab:Bert m2m}
\centering
\begin{tabular}{lllll}
\hline
    
  & \multicolumn{2}{c}{\textbf{M2M}}
  & \multicolumn{2}{c}{\textbf{MultiWOZ}}
\\
    \textbf{Model}
  & \textbf{ID Acc} 
  & \textbf{SF F1}
  & \textbf{ID Acc} 
  & \textbf{SF F1}
\\ \hline

$BERT_{base}$         
& 86.75 & 85.43 
& 65.34 & 92.18 
\\

$BERT_{large}$        
& 87.53 & 89.23 
& 72.49 & 95.32 
\\

$ALBERT_{base}$       
& 86.54 & 85.42 
& 65.31 & 91.87 
\\

$ALBERT_{large}$      
& 86.85 & 86.51 
& 72.46 & 93.41 
\\ \hline

$BERT_{base}$SA    
& 92.36 & 88.54 
& 69.76 & 95.81 
\\

$BERT_{large}$SA   
& 94.06 & 90.56 
& 75.01 & 97.74 
\\

$ALBERT_{base}$SA  
& 90.43 & 88.14 
& 70.24 & 96.32 
\\

$ALBERT_{large}$SA 
& 94.06 & 89.44 
& 74.31 & 97.41 
\\ \hline

\textbf{Our model}
  & \textbf{94.19}
  & \textbf{93.02} 
  & \textbf{78.49}
  & \textbf{97.98}
 \\ 

\textbf{with $\textbf{BERT}_{\textbf{base}}$}

  & $\pm0.23$
  & $\pm0.09$
  & $\pm0.29$
  & $\pm0.12$
  \\ 

  \hline
\end{tabular}
\end{table}

Table \ref{tab:joint multiwoz} shows the evaluation results compared to joint NLU baseline models. Different to the baselines, our model includes BERT embedding, the domain task, and explicit task interaction in each direction between all 3 tasks. As seen in Table \ref{tab:Bert m2m} the explicit interaction factor's contribution is significant and it is again the case here, as only the slot gated model has explicit interaction and only in the intent2slot direction.
The outperformance is much larger on the MultiWOZ dataset. We propose that with the larger number of domains in this dataset that the inclusion of a domain task is of greater benefit to the other tasks.

\begin{table}
\caption{Comparison results with joint NLU models}
\label{tab:joint multiwoz}
\centering

\begin{tabularx}{0.47\textwidth}{lcccc}
\hline 
& \multicolumn{2}{c}{\textbf{M2M}}
& \multicolumn{2}{c}{\textbf{MultiWOZ}}

\\ 
\textbf{Model}
& \textbf{ID Acc} 
& \textbf{SF F1} 
& \textbf{ID Acc} 
& \textbf{SF F1} 
\\ 

\hline
Bi-LSTM + CRF 
& 89.40
& 90.09
& 69.55     & 88.75     
           
\\
Seq2Seq \cite{liu2016attention}        
& 92.50
& 91.72
& 66.41     & 85.43                     
\\
Slot-Gated \cite{goo2018slot}  
& 93.27
& 92.79
& 68.83     & 87.76   
                     
\\ \hline
\textbf{Our model} 
& \textbf{94.19}
& \textbf{93.02}
& \textbf{78.49}
& \textbf{97.98}
\\ 

$\textbf{with BERT}_{\textbf{base}}$
& $\pm0.23$
& $\pm0.09$
& $\pm0.29$
& $\pm0.12$
\\ 
\hline
\end{tabularx}
\end{table}


\subsection{Effect of the Domain Classification Task}
To investigate the effect of inclusion of each of the sub-tasks we ran our model with only SF, only ID, then in all pairs from \{SF, ID, DC\}, and then with all three sub-tasks. As illustrated in Table \ref{tab:ablation:joint}, the inclusion of a domain task improves the result of the same experiment without such a task in each case, except SF to SF+DC for M2M.
Including DC with an ID task gives better improvement than including ID with SF, indicating that the domain guides the intents present more than the slots present. Then the inclusion of DC with ID and SF significantly improves the SF performance, as DC improves ID, which then flows down the semantic frame to a better SF result.

\begin{table}
\caption{Comparison results with different NLU tasks (ID - Intent Detection, SF - Slot Filling, and DC - Domain Classification) performed}
\label{tab:ablation:joint}
\centering
\begin{tabularx}{0.47\textwidth}{lcccc}
\hline
& \multicolumn{2}{c}{\textbf{M2M}}
& \multicolumn{2}{c}{\textbf{MultiWOZ}}

\\ 
\textbf{NLU Tasks}
& \textbf{ID Acc} 
& \textbf{SF F1}
& \textbf{ID Acc} 
& \textbf{SF F1} 
\\ 
\hline
SF   
& -
& 90.20
& -      & 86.53 
\\
ID 
& 91.39
& -
& 73.38 & -         
\\
\text{SF + ID}
& 93.73
& 91.42
& 77.91 & 92.43 
\\
\text{SF + DC}  
& -
& 90.15
& -      & 89.92 
\\
\text{ID + DC} 
& 93.04
& -
& 78.11 & -          
\\ \hline
\text{SF + ID + DC} 
& \textbf{94.19}
& \textbf{93.02}
& \textbf{78.49}
& \textbf{97.98}
\\ 
& $\pm0.23$
& $\pm0.09$
& $\pm0.29$
& $\pm0.12$
\\ 
\hline
\end{tabularx}
\end{table}

\subsection{Effect of Transformer Encoder Interaction}
We tested the three tri-Transformer-encoder variants. As illustrated in Table \ref{tab:ablation:layer}, ID performance benefits from some interaction at this level; for M2M it is ambivalent to Cross-Attention or BF, while MultiWOZ prefers Cross-Attention.
This indicates that the sharing of slot information with intent before self-attention is more effective for ID than performing further self-attention on the slot information and sharing the results. For SF though, M2M gets a benefit from Cross-Attention but BF performs worse than NoEx, while for MultiWOZ Cross-Attention is worse than NoEx but there is a marked benefit from BF exchange. This indicates that, in MultiWOZ, SF benefits from extra SA only on its own Q, K and V vectors, for the purpose of extracting context information, before incorporating domain and intent information in its FF layer. The longer turn length of this dataset, and the multi-intent nature of some turns (albeit joined into a single intent label), may contribute to this outcome.

\begin{table}
\caption{Comparison results with transformer encoder variants}
\label{tab:ablation:layer}
\centering
\begin{tabularx}{0.47\textwidth}{lcccc}
\hline
&  
\multicolumn{2}{c}{\textbf{M2M}}
&  
\multicolumn{2}{c}{\textbf{MultiWOZ}}
\\ 
\textbf{Encoder Variant} 
& \textbf{ID Acc}
& \textbf{SF F1} 
& \textbf{ID Acc}
& \textbf{SF F1} 
\\
\hline

NoEx  
& 82.38 & 92.69      
& 76.00          & 93.82        

\\
Cross 
& \textbf{94.19} & \textbf{93.02}      
& \textbf{79.83} & 90.92          
         
        
\\
BF    
& \textbf{94.19}
& 91.91
& 78.49          & \textbf{97.98} 

\\

\hline
\end{tabularx}
\end{table}

\section{Discussion and Conclusion} \label{chap:conclusion}
We move to datasets beyond the standard single-turn utterance benchmarks used in NLU. We propose a novel model in which slot, intent, and domain predictions explicitly exchange information in two model layers. The inclusion of the domain task improves performance in the slot and intent tasks. We achieved superior results on two multi-turn utterance datasets compared to baseline models and state-of-the-art joint models. By comparing our model to larger BERT models with more parameters, we can posit that the boosting effect is due to the structure of our model and can not be surpassed by larger BERT models. 
As far as we know, we are the first to conduct domain classification, slot filling and intent detection experiments on multi-turn datasets. While we utilise multi-turn datasets, we do not use slot, intent and domain information from previous dialogue turns. In future studies, we will incorporate such context. The different datasets give conflicting results on the type of explicit exchange between tasks that gives the best results. 


\bibliographystyle{IEEEtran}
\bibliography{references}


\end{document}